\documentclass[conference]{IEEEtran}
\IEEEoverridecommandlockouts

\usepackage{cite}
\usepackage{amsmath,amssymb,amsfonts}
\usepackage{algorithmic}
\usepackage{graphicx}
\usepackage{textcomp}
\usepackage{xcolor}
\usepackage{subfig}
\def\BibTeX{{\rm B\kern-.05em{\sc i\kern-.025em b}\kern-.08em
    T\kern-.1667em\lower.7ex\hbox{E}\kern-.125emX}}
\begin{document}

\title{Instance Camera Focus Prediction for Crystal Agglomeration Classification}

\author{
    \IEEEauthorblockN{Xiaoyu Ji\IEEEauthorrefmark{1}, Chenhao Zhang\IEEEauthorrefmark{1}, Tyler James Downard\IEEEauthorrefmark{2}, Zoltan Nagy\IEEEauthorrefmark{2}, Ali Shakouri\IEEEauthorrefmark{1}, Fengqing Zhu\IEEEauthorrefmark{1}}
    \IEEEauthorblockA{\IEEEauthorrefmark{1} Elmore School of Electrical and Computer Engineering, Purdue University,\\ West Lafayette, IN 47907, USA
    }
    \IEEEauthorblockA{\IEEEauthorrefmark{2}Davidson School of Chemical Engineering, Purdue University,\\ West Lafayette, IN 47907, USA
    }
}

\maketitle

\begin{abstract}
Agglomeration refers to the process of crystal clustering due to interparticle forces. Crystal agglomeration analysis from microscopic images is challenging due to the inherent limitations of two-dimensional imaging. Overlapping crystals may appear connected even when located at different depth layers. Because optical microscopes have a shallow depth of field, crystals that are in-focus and out-of-focus in the same image typically reside on different depth layers and do not constitute true agglomeration. To address this, we first quantified camera focus with an instance camera focus prediction network to predict 2-class focus level that aligns better with visual observations than traditional image processing focus measures. Then an instance segmentation model is combined with the predicted focus level for agglomeration classification. Our proposed method has a higher agglomeration classification and segmentation accuracy than the baseline models on ammonium perchlorate crystal and sugar crystal dataset.

\end{abstract}


\section{Introduction}

Crystallization is a fundamental process that forms solid crystals from particles. The degree of crystal agglomeration is an important metric in this process that influences the stability and size distribution of the resulting solid products \cite{Hui2024,Brun05,Alander04}. Agglomeration refers to the clustering of particles, where bridges form between them and trap water. To analyze this behavior, microscopic image analysis has been widely used in this scenario \cite{ho17,Timothy25,ji24}.

In microscopic imaging, liquid samples containing crystals suspended in solvent are captured as two-dimensional images, making crystal agglomeration classification a challenging task. Although some crystals appear to be attached, they may actually reside on different depth layers and are not physically connected. Previous single-view agglomeration classification works have reported high performance using multi-class instance segmentation deep learning models \cite{Jiang21,Timothy25}, typically operating under the assumption that all crystals reside within the same depth layer. These approaches do not consider the effects of camera focus, which can lead to misclassification of crystals located at different focal planes.

\begin{figure}[t]
  \centering
  \includegraphics[scale=0.6]{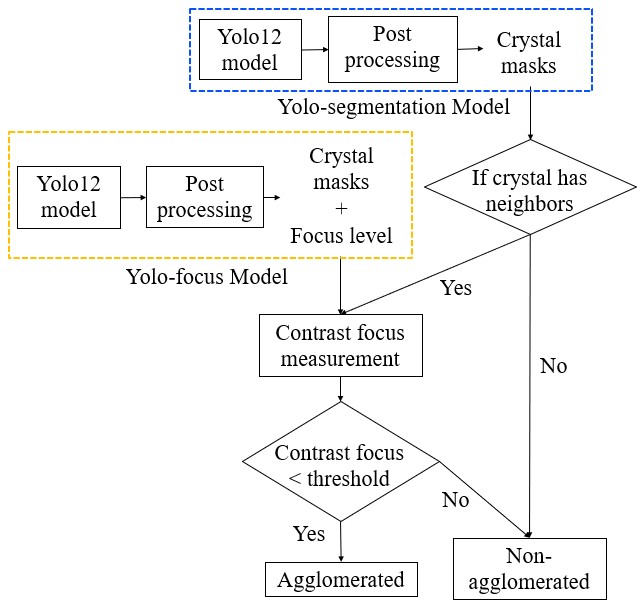}
  \caption{Overview of our proposed agglomeration classification method. Yolo-Segmentation model predicts instance segmentation masks and the agglomeration classification is based on Yolo-focus model predictions and contrast focus measurement.}
  \label{fig:1}
\end{figure}

Camera focus can be estimated using traditional image processing focus measure techniques because the degree of blurriness is proportional to the amount of defocus, such as the Laplacian variance \cite{Sub1993}, Brenner function \cite{Brenner1976}, and Reblur function \cite{crete07}. However, our experiments show that these objective functions perform poorly when applied to instance-masked regions within an image. Depth estimation provides an alternative approach to infer camera focus, since focus level is correlated with depth. Yet, pretrained depth estimation models on in-door or out-door scenes \cite{depth_anything_v2} do not generalize well to microscopic images, as validated in our experiments. Moreover, finetuning a depth estimation model on crystal data is challenging because acquiring accurate ground-truth depth information for transparent microscopic objects is inherently difficult.

Given the challenges of estimating instance-wise camera focus levels in microscopic images, we propose a supervised learning–based camera focus model. The training dataset is manually annotated by an domain expert, who assigns focus levels (in-focus or out-of-focus) to individual crystals based on visual inspection. The model is based on Yolov12 \cite{tian2025yolov12} instance segmentation model, named as Yolo-focus. The predicted focus levels are used to compute the contrast focus—defined as the focus level difference between a crystal and its neighboring instances—which serves as a key metric for distinguishing agglomerated from non-agglomerated crystals within a cluster.

Our proposed method integrates both the Yolo-focus model and a Yolo based segmentation model (named as Yolo-segmentation), as illustrated in Figure \ref{fig:1}. The final instance masks are generated by Yolo-segmentation model, while their agglomeration classification are based on the contrast focus values derived from the predictions of Yolo-focus model. Compared with the results obtained using baseline models (Mask Region-based Convolutional Neural Network (Mask R-CNN) \cite{He17} and Yolov12 \cite{tian2025yolov12}) and ablation studies (Yolo-focus model only), our method achieves the highest agglomeration classification and segmentation accuracy. Visual assessments further demonstrate the effectiveness of our approach, particularly in handling crystals at different focus levels.
\section{Method}
We propose a camera focus prediction based agglomeration classification method. As shown in Figure \ref{fig:1}, Yolo-segmentation model and Yolo-focus model are integrated with contrast focus measurement.

\subsection{Yolo-segmentation Model}
Yolo-segmentation model is based on the YOLO12 instance segmentation framework \cite{tian2025yolov12}. In our method, we use only the segmentation outputs of this model. However, it is trained with agglomeration class labels to improve segmentation accuracy. This model also serves as a Yolov12 baseline comparison method, as discussed in section \ref{exp}. Post processing shown in Figure \ref{fig:1} includes instance mask hole infilling and largest connected component preserving to ensure the enclosing property of the crystals.

\subsection{Yolo-focus Model}

The Yolo-focus model is built upon the YOLO12 instance segmentation framework \cite{tian2025yolov12}. The model separates crystal instances into two classes (focus levels): in-focus and out-of-focus. During training, we modified the YOLO12 dataloader by disabling mosaic, mixup, and copypaste augmentations. These augmentations were excluded because the ground-truth focus-level labels depend on the intrinsic focus distribution of each individual image. When multiple images are composited through such augmentations, the focus distribution is altered while the associated focus-level labels remain unchanged, potentially leading to label inconsistency.

An additional data augmentation technique, instance blurring, was introduced to the training dataset. In this process, a Gaussian blur is randomly applied to the crystal instances. The blurred instances are labeled as out-of-focus, thereby enhancing the dataset’s diversity and improving the model’s robustness to defocus variations.

\subsection{Contrast Focus Measurement}

In Figure \ref{fig:1}, we performed contrast focus measurement on the output of the Yolo-focus model prior to thresholding for classification. Contrast focus is defined as the deviation of an instance’s predicted focus level from the focus levels of its neighboring instances. A higher contrast focus indicates a greater depth difference between crystals, suggesting that the crystals lie on distinct depth layers and are therefore not physically agglomerated. Based on this assumption, we evaluated two methods for contrast focus measurement:
\begin{itemize}
\item \textit{Contrast 1}: The difference between the instance’s predicted focus level and the mean focus level of its neighboring instances.

\item \textit{Contrast 2}: Assigning a low contrast focus when any of its neighboring instances has the same focus level as the instance itself, and a high contrast focus otherwise.
\end{itemize}
The \textit{Contrast 2} method demonstrated superior performance and was adopted as the final approach. Detail comparisons are illustrated in section \ref{exp}.

\begin{table*}[htbp]
\caption{Agglomeration classification and segmentation metrics tested on \textit{Dataset 1}. All the metrics are the higher the better.}
\centering  
\label{tab:table_1}
\normalsize
\begin{tabular}{|p{0.55\columnwidth}|p{0.15\columnwidth}|p{0.18\columnwidth}|p{0.18\columnwidth}|p{0.18\columnwidth}|p{0.18\columnwidth}|} 
\hline
Methods & ACC (\%)&  F1 (\%)& IoU (\%) & AP (\%) & Recall (\%) \\
\hline
Mask R-CNN \cite{He17} & 66.307 & 63.123 & 57.206 & 75.073 & 55.476\\
\hline
Yolov12 \cite{tian2025yolov12} & 80.904 & 76.641 & 63.019 & 75.575 & 78.039\\
\hline\hline
Our ablation (\textit{Contrast 1}) & 82.809 & 76.875 & 63.426 & 76.582 & 77.527\\
\hline
Our ablation (\textit{Contrast 2}) & 82.276 & 76.485 & 63.053 & 76.298 & 77.009\\
\hline
Ours (\textit{Contrast 1}) &  85.421 & 80.629 & 66.381 & 79.533 & 82.112\\
\hline
Ours (\textit{Contrast 2}) &  \textbf{85.503} & \textbf{80.777} & \textbf{66.407} & \textbf{79.677} & \textbf{82.266}\\
\hline
\end{tabular}
\end{table*} 

\begin{table}[htbp]
\caption{Agglomeration classification accuracies tested on \textit{Dataset 2}. Methods from left to right are two baseline models, our ablation model with \textit{Contrast 1} and our method with \textit{Contrast 2}.}
\centering  
\label{tab:table_2}
\normalsize
\begin{tabular}{|p{0.15\columnwidth}|p{0.18\columnwidth}|p{0.12\columnwidth}|p{0.2\columnwidth}|p{0.1\columnwidth}|} 
\hline
Methods & Mask R-CNN \cite{He17}&  Yolov12 \cite{tian2025yolov12}& Our ablation & Ours \\
\hline
ACC (\%) & 53.120 & 53.832 & 64.947 & \textbf{68.310}\\
\hline
\end{tabular}
\end{table}

\section{Dataset}

The training datasets for Yolo-segmentation model and Yolo-focus model are derived from the same set of images but use different labeling schemes. A total of 11 original microscopic images of a fixed size $1224\times920$ were collected. Segmentation masks, agglomeration classification labels and focus level labels were first manually annotated by an domain expert, after which instance blurring data augmentation was applied, expanding the dataset to 55 images. The Gaussian blur was randomly applied to 25–50\% of the crystal instances, with the kernel size randomly selected between 11 and 17. Blurred instances in the augmented images are labeled as out-of-focus. 

There are two test datasets: \textit{Dataset 1} is ammonium perchlorate crystal dataset includes 25 images of the same size as training data; \textit{Dataset 2} is sugar crystal dataset includes 10 images with size $256\times 256$. Both datasets include manually labeled masks and agglomeration class labels by domain experts.

The manual annotation criterion for agglomeration class includes: 1) In-focus and out-of-focus objects located on different depth layers are not considered agglomerated, even if they are touching; 2) Instances that are both in focus and in edge contact are considered agglomerated; 3) Small polygonal shapes positioned on top of larger ones may represent homogeneous breakages and are not labeled as agglomeration; 3) Small, solid black circular objects without internal openings are regarded as dust or bubbles and are excluded from agglomeration labeling.

\section{Experiments}
\label{exp}

\begin{figure}[ht]
  \centering
  \includegraphics[scale=0.55]{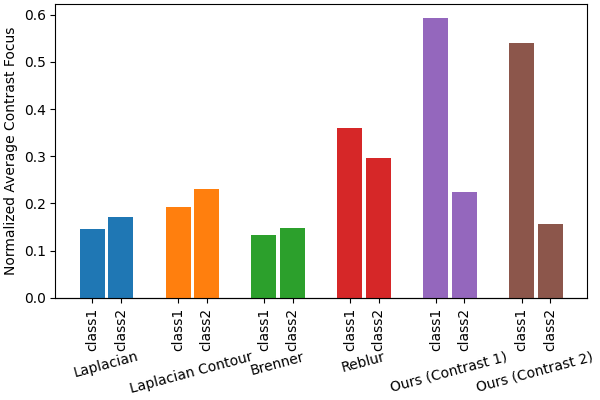}
  \caption{Normalized average contrast focus comparison. Class1 and class2 represent non-agglomerated and agglomerated crystals. The normalized average contrast focus is calculated as the mean contrast focus of each method divided by the maximum contrast focus. From left to right are  Laplacian variance \cite{Sub1993} on entire mask and mask contour, Brenner \cite{Brenner1976}, Reblur \cite{crete07} and our methods. A larger difference between class 1 and class 2 indicates better separation of focus levels.}
  \label{fig:2}
\end{figure}

\subsection{Experimental setting}
For our proposed method, we initialized Yolo-segmentation model using the pre-trained weights from the COCO dataset \cite{coco14}, while Yolo-focus model was trained from scratch without using any pre-trained weights. Both models were trained for a maximum of 1,000 epochs, employing early stopping with a tolerance of 100 epochs. The training processes were performed on two NVIDIA GeForce GTX 1080 Ti GPUs. The threshold for contrast focus is set as 0.5.

We set ablation studies to compare our proposed method with a baseline configuration that used only Yolo-focus model combined with contrast focus measurement. The ablation methods differ from the proposed approach in that the predicted instance masks are generated by Yolo-focus model rather than Yolo-segmentation model. Also, the Yolov12 baseline results shown in the third row of Table \ref{tab:table_1} are the results of Yolo-segmentation model.


\subsection{Quantitative Comparison}
\subsubsection{Agglomeration Classification and Segmentation Accuracy}

In Table \ref{tab:table_1}, we evaluated two baseline models, two ablation models and our methods with agglomeration classification accuracy metric (ACC) and four segmentation accuracy metrics on \textit{Dataset 1}. F1 and intersection-over-union (IoU) are general instance-wise and pixel-wise segmentation metrics. Average precision (AP) and Recall are instance-wise metrics measuring the predicted instance accurate ratio and the ground truth instance accurate ratio. F1, AP and Recall used an IoU confidence threshold of 50\%.

Based on the quantitative results, our ablation models achieved higher accuracy across most evaluation metrics compared to the baseline models. The ablation model with \textit{Contrast 1} method performs better than that with \textit{Contrast 2} method. Compared to that, our proposed method further improves both the agglomeration classification accuracy and segmentation accuracy, where \textit{Contrast 2} method has a better performance than \textit{Contrast 1} method. 

The agglomeration classification of \textit{Dataset 2} is shown in Table \ref{tab:table_2}. Our proposed method also has a higher accuracy than other methods.

\subsubsection{Instance Camera Focus Distribution}

To effectively distinguish in-focus from out-of-focus instances, we expect that crystals (with neighbors) labeled as agglomerated in the ground truth will exhibit a different instance-level camera focus distribution compared to those labeled as non-agglomerated. The predicted average contrast focus of our proposed Yolo-focus model is compared against traditional image processing focus measures, including the Laplacian variance \cite{Sub1993}, Brenner \cite{Brenner1976}, and Reblur \cite{crete07} objective functions.

Figure \ref{fig:2} presents the normalized average contrast focus of different methods on test \textit{Dataset 1}. The bar plot shows the comparison between two classes: non-agglomerated crystal and agglomerated crystal. There are 201 agglomerated crystals and 68 non-agglomerated crystals. Traditional focus measures show minimal separation between the normalized average contrast focus of the two classes, while our method achieves a significantly larger difference, indicating clearer class distinction. The two contrast focus measurement settings perform similarly under our approach.

\subsection{Qualitative Comparison}

\begin{figure}
    \centering
  \subfloat[original image\label{3a}]{%
       \includegraphics[width=0.3\linewidth]{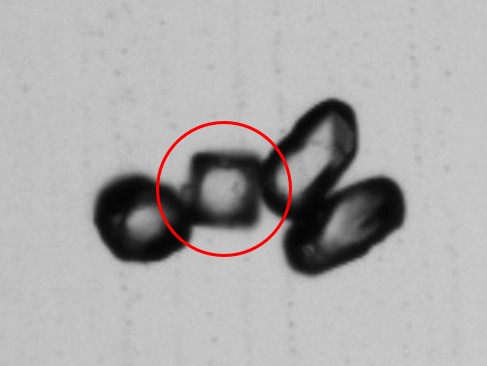}}
    \hfill
    \subfloat[original image\label{3b}]{%
       \includegraphics[width=0.3\linewidth]{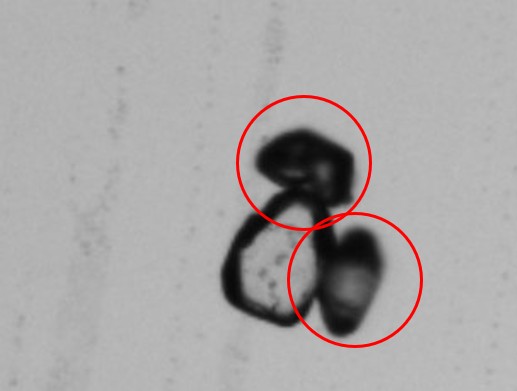}}
    \hfill
  \subfloat[original image\label{3c}]{%
        \includegraphics[width=0.3\linewidth]{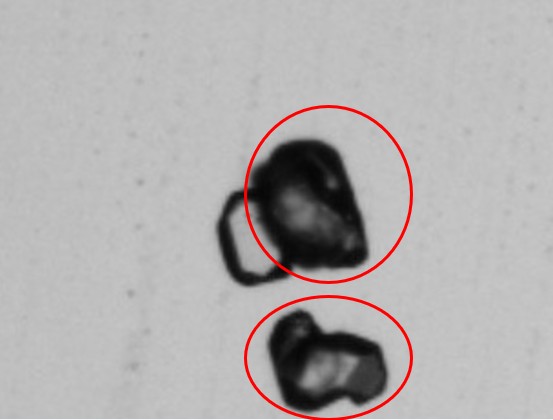}}
   \\
  \subfloat[Yolov12 \cite{tian2025yolov12} \label{3d}]{%
        \includegraphics[width=0.3\linewidth]{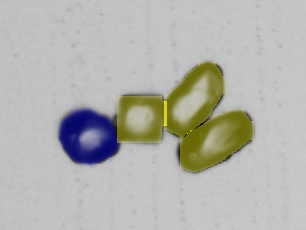}}
     \hfill
  \subfloat[Yolov12 \cite{tian2025yolov12} \label{3e}]{%
        \includegraphics[width=0.3\linewidth]{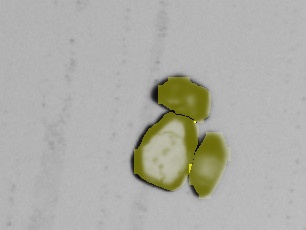}}
    \hfill
  \subfloat[Yolov12 \cite{tian2025yolov12} \label{3f}]{%
        \includegraphics[width=0.3\linewidth]{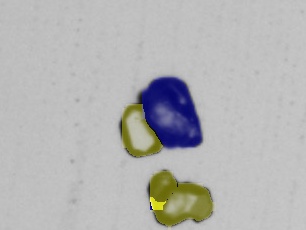}}
    \\
  \subfloat[Mask R-CNN \cite{He17} \label{3g}]{%
        \includegraphics[width=0.3\linewidth]{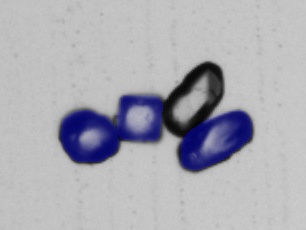}}
     \hfill
  \subfloat[Mask R-CNN \cite{He17} \label{3h}]{%
        \includegraphics[width=0.3\linewidth]{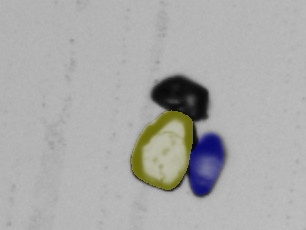}}
    \hfill
  \subfloat[Mask R-CNN \cite{He17} \label{3i}]{%
        \includegraphics[width=0.3\linewidth]{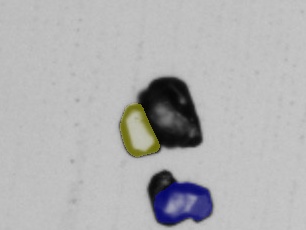}}
    \\
  \subfloat[Ours \label{3j}]{%
        \includegraphics[width=0.3\linewidth]{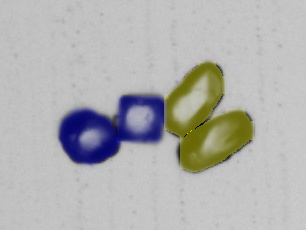}}
     \hfill
  \subfloat[Ours \label{3k}]{%
        \includegraphics[width=0.3\linewidth]{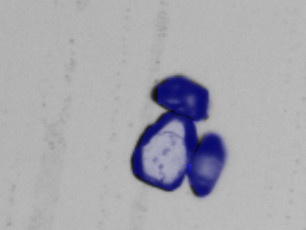}}
    \hfill
  \subfloat[Ours \label{3l}]{%
        \includegraphics[width=0.3\linewidth]{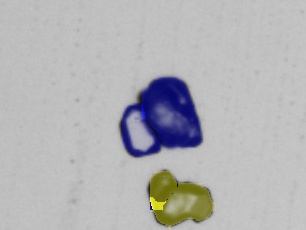}}
        \\
  \subfloat[GT \label{3m}]{%
        \includegraphics[width=0.3\linewidth]{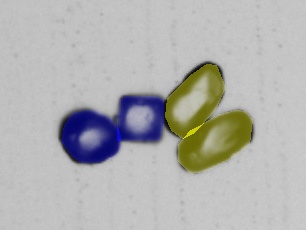}}
     \hfill
  \subfloat[GT \label{3n}]{%
        \includegraphics[width=0.3\linewidth]{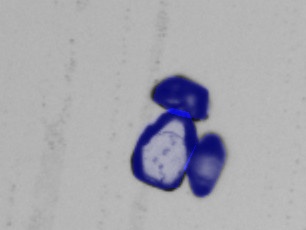}}
    \hfill
  \subfloat[GT \label{3o}]{%
        \includegraphics[width=0.3\linewidth]{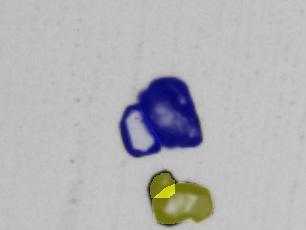}}
        \\
  \subfloat[Predicted Out-of-focus mask \label{3p}]{%
        \includegraphics[width=0.3\linewidth]{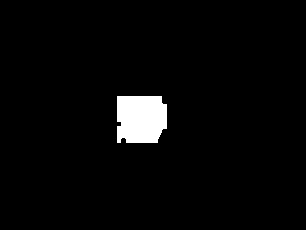}}
     \hfill
  \subfloat[Predicted Out-of-focus mask \label{3q}]{%
        \includegraphics[width=0.3\linewidth]{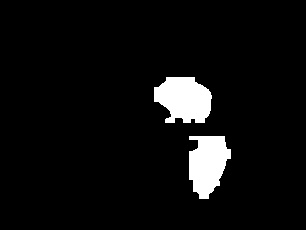}}
    \hfill
  \subfloat[Predicted Out-of-focus mask \label{3r}]{%
        \includegraphics[width=0.3\linewidth]{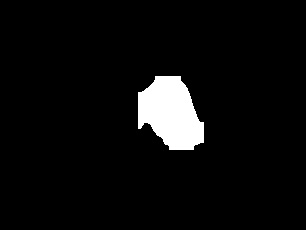}}
  \caption{Visual comparison of three example images. Red circles highlight out-of-focus crystals in (a)-(c). Blue color indicates non-agglomerated and yellow color indicates agglomerated.}
  \label{fig:3} 
\end{figure}

As shown in Figure \ref{fig:3}, we compare the predicted agglomeration classification masks obtained from different methods across three example images. In each image, there are crystals appear clustered but are labeled as non-agglomerated due to high contrast focus levels. The red-circles in subfigures \ref{3a}–\ref{3c} highlight relatively out-of-focus crystals, which are assumed to lie on different depth layers from their neighboring crystals. As shown in the ground truth masks in subfigures \ref{3m}–\ref{3o}, only crystals that are attached and share the same focus level are classified as agglomerated (yellow), while all others are labeled as non-agglomerated (blue).

The baseline predicted masks shown in the subfigures \ref{3d}-\ref{3i} exhibit cases of absent crystals or misclassified crystals in clusters. While our proposed method yields accurate classifications in subfigures \ref{3j}-\ref{3l}. This is because the classification is derived from the estimated focus levels. The subfigures \ref{3p}-\ref{3r} display the out-of-focus masks predicted by Yolo-focus model. Except for the mask in subfigure \ref{3r} of example 3, which misses two crystals, all predicted masks are accurate. Nevertheless, this minor omission does not affect the overall classification accuracy, as the two missed crystals are both in-focus and thus belong to the same focus level.

\section{Conclusion}

In this paper, we propose a camera focus model for agglomeration classification in microscopic crystal images. The contrast in focus between a crystal and its neighboring crystals serves as a key metric for distinguishing agglomerated from non-agglomerated crystals. By analyzing contrast focus distributions, we confirmed that the trained camera focus model enhances the distinction between class distributions. Our integrated approach—combining the camera focus model with the segmentation model—achieves superior performance compared to baseline methods.

\bibliographystyle{IEEEtran}
\bibliography{main}

\end{document}